\documentclass[sigconf]{acmart}

\usepackage{algorithm}
\usepackage{algorithmic}
\usepackage{xcolor}
\usepackage{multirow}
\usepackage{booktabs}
\usepackage{multicol}
\usepackage{makecell}
\usepackage{algorithmic}
\usepackage{graphicx}

\usepackage{cleveref}
\usepackage{nicefrac}
\usepackage{siunitx}
\usepackage{relsize}
\usepackage{amsthm}

\usepackage{booktabs}
\usepackage{colortbl} 
\usepackage{graphicx} 
\usepackage[table]{xcolor} 
\usepackage{xspace}
\usepackage{enumitem}

\usepackage{subcaption}
\newcommand{\T}{FlashFPS\xspace}

\BeforeBeginEnvironment{equation*}{\begingroup\vspace{-5pt}}
\AfterEndEnvironment{equation*}
{\endgroup}

\AtBeginDocument{%
  }

\setcopyright{acmlicensed}

\copyrightyear{2026}
\acmYear{2026}
\setcopyright{cc}
\setcctype{by}
\acmConference[DAC '26]{63rd ACM/IEEE Design Automation Conference}{July 26--29, 2026}{Long Beach, CA, USA}
\acmBooktitle{63rd ACM/IEEE Design Automation Conference (DAC '26), July 26--29, 2026, Long Beach, CA, USA}
\acmDOI{10.1145/3770743.3803907}
\acmISBN{979-8-4007-2254-7/2026/07}

\begin{document}
\title{FlashFPS: Efficient Farthest Point Sampling for Large-Scale\\ Point Clouds via Pruning and Caching}


\author{Yuzhe Fu, Hancheng Ye, Cong Guo$\dagger$, Junyao Zhang, Qinsi Wang, Yueqian Lin, Changchun Zhou$\dagger$, Hai (Helen) Li and Yiran Chen}
    \affiliation{%
      \institution{Duke University}
      \city{Durham}
      \state{North Carolina}
      \country{USA}
    }
\email{{yuzhe.fu, cong.guo, changchun.zhou, hai.li, yiran.chen}@duke.edu}

\renewcommand{\shortauthors}{Y. Fu, H. Ye, C. Guo, J. Zhang, Q. Wang, Y. Lin, C. Zhou, H. Li, Y. Chen}

\begin{abstract}
Point-based Neural Networks (PNNs) have become a key approach for point cloud processing. 
However, a core operation in these models, Farthest Point Sampling (FPS), often introduces significant inference latency, especially for large-scale processing. 
Despite existing CUDA- and hardware-level optimizations, FPS remains a major bottleneck due to exhaustive computations across multiple network layers in PNNs, which hinders scalability.
Through systematic analysis, we identify three substantial redundancies in FPS, including unnecessary full-cloud computations, redundant late-stage iterations, and predictable inter-layer outputs that make later FPS computations avoidable.
To address these, we propose \textbf{\textit{FlashFPS}}, a hardware-agnostic, plug-and-play framework for FPS acceleration, composed of \textit{FPS-Prune} and \textit{FPS-Cache}. \textit{FPS-Prune} introduces candidate pruning and iteration pruning to reduce redundant computations in FPS while preserving sampling quality, and \textit{FPS-Cache} eliminates layer-wise redundancy via cache-and-reuse. Integrated into existing CUDA libraries and state-of-the-art PNN accelerators, \textit{FlashFPS} achieves 5.16× speedup over the standard CUDA baseline on GPU and 2.69× on PNN accelerators, with negligible accuracy loss, enabling efficient and scalable PNN inference. Codes are released at \url{https://github.com/Yuzhe-Fu/FlashFPS}.

\end{abstract}

\maketitle

\section{Introduction}

\footnotetext[0]{$\dagger$ Corresponding Author: Cong Guo and Changchun Zhou}

\begin{figure}[t]
    \centering
    \includegraphics[width=0.95\linewidth]{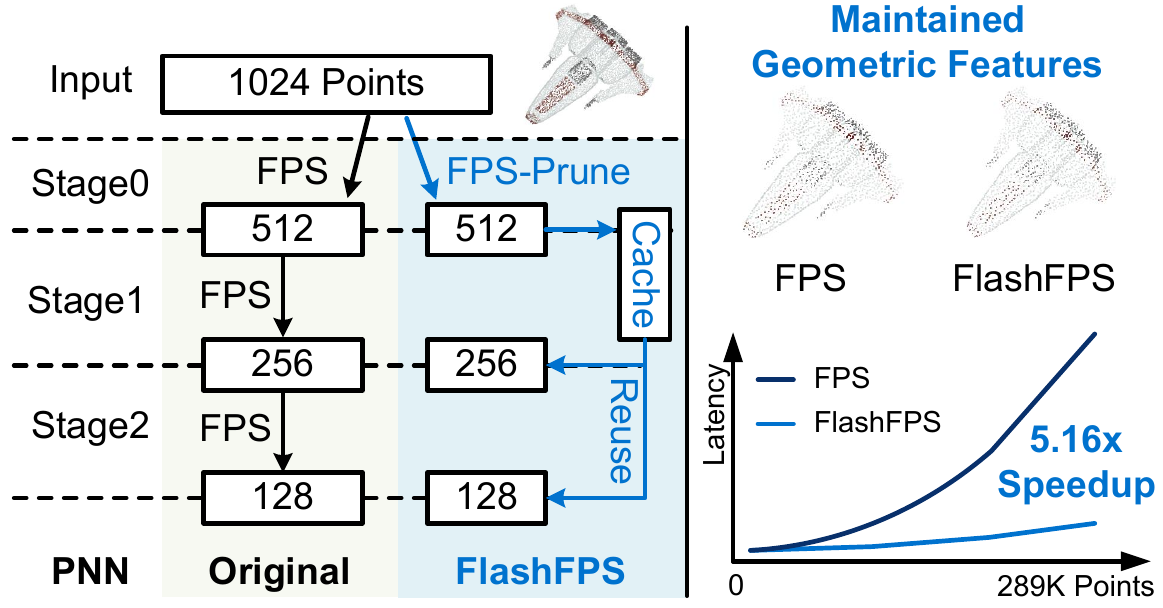}
    \caption{(Left) Comparison between the original FPS-based (CUDA-optimized) and FlashFPS-based PNN architecture. (Right) FlashFPS well maintains geometric features and achieves an average 5.16x inference speedup for PNNs.}
    \label{fig:intro}
\end{figure}

Point-based Neural Networks (PNNs) have become the foundation for 3D perception in applications such as robotics~\cite{zhu2024point}, autonomous driving~\cite{guan2025talk2pc}, and AR/VR systems~\cite{li2024fusion}. PNNs typically adopt a hierarchical architecture~\cite{qi2017pointnet, qi2017pointnet++}, as shown in Fig.~\ref{fig:intro}, where each stage downsamples the point cloud via Farthest Point Sampling (FPS) and performs local feature aggregation. FPS has thus become a standard component in modern PNNs~\cite{he2025pointrwkv,deng2023pointvector,qian2022pointnext}, valued for its ability to maintain geometric coverage and spatial uniformity.

\begin{figure}[tb]
    \centering
    \includegraphics[width=0.9\linewidth]{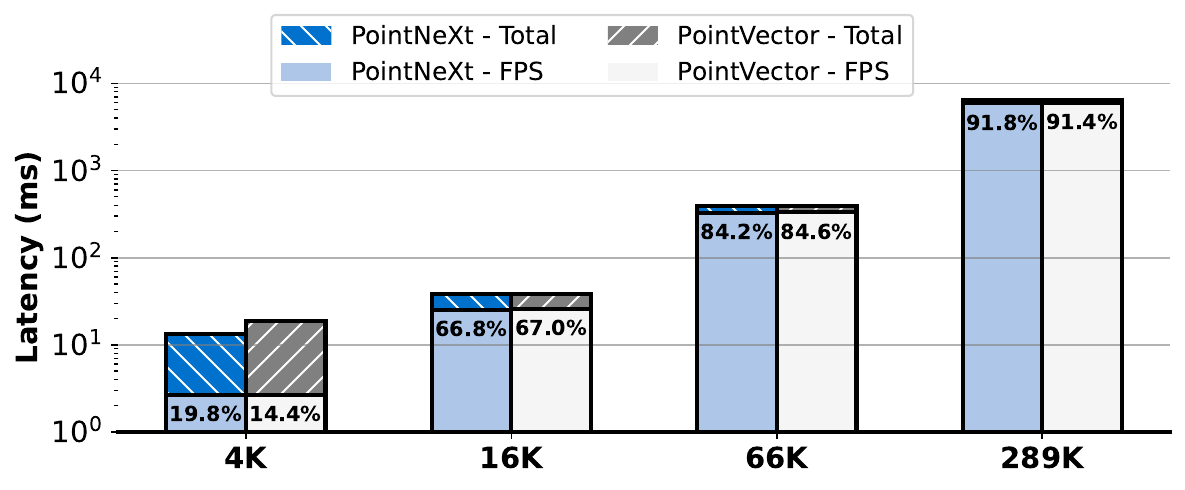}
    \caption{The latency breakdown for PointNeXt-L and PointVector-L inference on a GPU for S3DIS segmentation workloads under varying input scales.}
    \label{fig:latency}
\end{figure}

\begin{figure}[tb]
    \centering
    \includegraphics[width=0.75\linewidth]{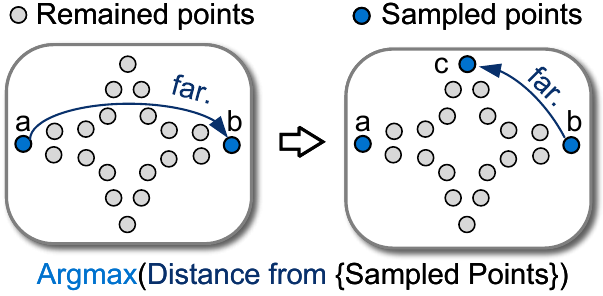}
    \caption{Illustration of iteratively greedy selection in FPS.}
    \label{fig:FPS}
\end{figure}

However, as PNNs scale toward real-world deployment, the number of input points has increased drastically, from 1K in early benchmarks~\cite{modelnet40} to over 200K in semantic segmentation datasets~\cite{armeni20163d, dai2017scannet}, and around 300K per frame in real LiDAR applications~\cite{roriz2024survey}. 
This rapid growth leads to quadratically increased inference latency. 
Profiling results show that FPS alone contributes over 90\% of the total runtime on GPUs for large-scale point clouds, as shown in Fig.~\ref{fig:latency}. The inefficiency stems from the iterative greedy nature of FPS (Fig.~\ref{fig:FPS}): at each iteration, it computes the distances from all input points to the current sampled set and selects the farthest unsampled point as the next sample (highlighted in blue in Fig.~\ref{fig:latency}).
In large-scale processing scenarios, FPS remains essential, as simpler alternatives (e.g., random sampling) cause a 3\%–5\% accuracy drop that retraining cannot recover.
As a result, when input size grows, this exhaustive computation becomes prohibitive, making FPS the dominant bottleneck in modern PNN pipelines.

Prior FPS acceleration methods fall into two categories: CUDA-based and hardware-specific. The standard CUDA approaches~\cite{openpoints2023, qi2017pointnet++} improve intra-iteration parallelism but retain exhaustive computations and fully greedy loops.
QuickFPS~\cite{han2023quickfps} further improves CUDA efficiency by leveraging a kd-tree to organize memory, but its construction overhead outweighs the benefits when the input size is below 30K points.
Hardware-specific works~\cite{fu2026fractalcloud,gao2026lpcn,zhou202523,yoon2023efficient,feng2022crescent} mainly focus on memory streaming and reducing exhaustive operations via partitioning or prediction, but they demand dedicated hardware accelerators, often leading to low GPU utilization and even accuracy degradation, thus limiting scalability and flexibility.

Furthermore, a common limitation of existing optimizations is their neglect of the intrinsic algorithmic redundancy in FPS, redundancy that worsens as point clouds grow. While FPS ensures geometric coverage, its purely distance-based criterion is misaligned with the semantic objectives of 3D understanding tasks. This mismatch causes FPS to oversample semantically similar regions. For example, selecting both tips of an airplane wing due to geometric distance, despite their semantic redundancy. Such inefficiency becomes more pronounced at scale. Internally, this redundancy manifests in  two aspects: 
\textbf{(1) Exhaustive full-cloud computation across all points; (2) Repeated iterations inherent to the algorithm,} both of which contribute to excessive computation.

\begin{figure}[t]
    \centering
    \begin{subfigure}[t]{0.49\linewidth}
        \centering
        \includegraphics[width=\linewidth]{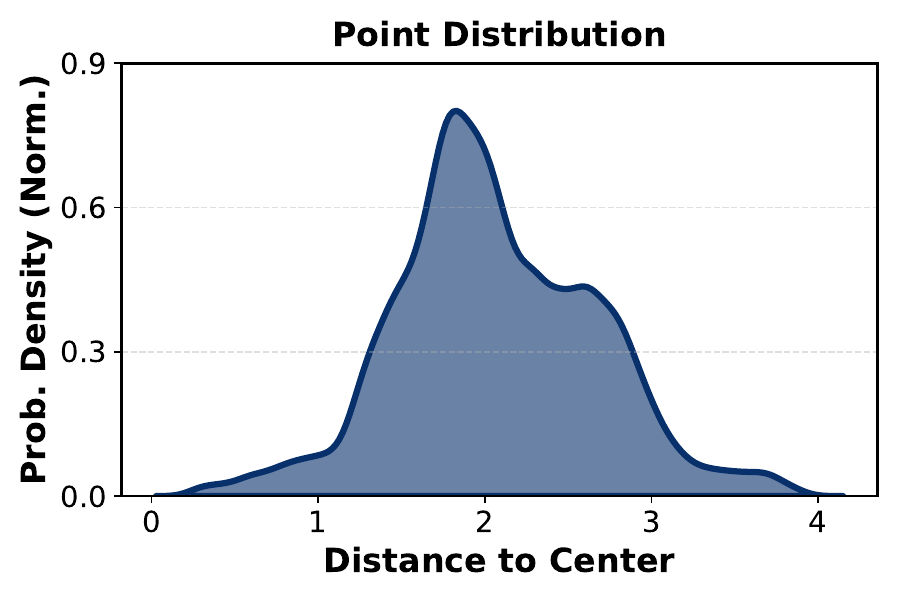}
        \caption{Original.}
        \label{fig:pointcloud_xy_kde_original}
    \end{subfigure}
    \hfill
    \begin{subfigure}[t]{0.49\linewidth}
        \centering
        \includegraphics[width=\linewidth]{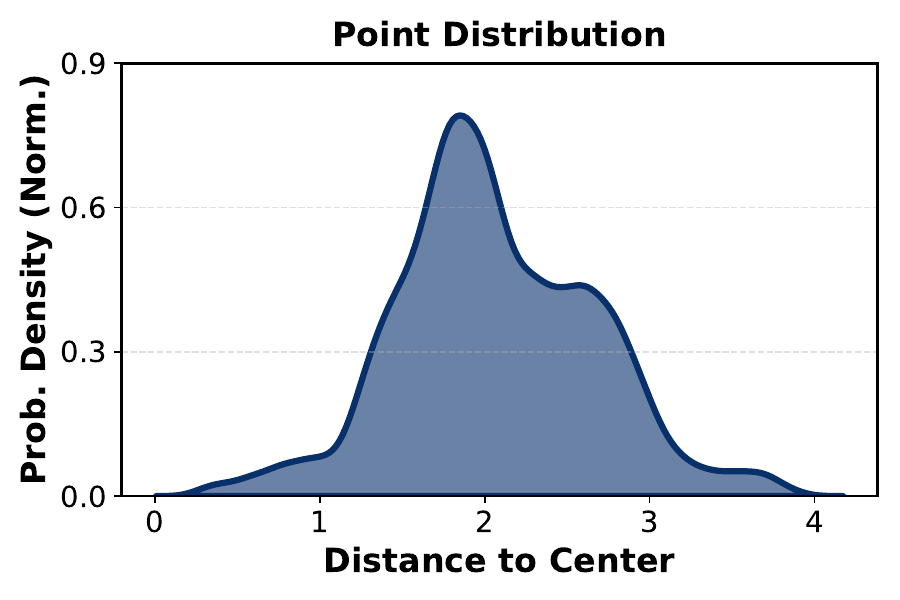}
        \caption{50\% random sparsification.}
        \label{fig:pointcloud_xy_kde_sparse}
    \end{subfigure}
    \caption{Point distributions after FPS for (a) the original and (b) sparsified input. The X-axis is the distance to the point cloud center and the Y-axis is the normalized probability density. }
    \label{fig:pointcloud_distribution_comparison}
\end{figure}

\vspace{5pt}
\noindent\textbf{Redundancy in Full-Cloud Computations.} 
FPS selects the point farthest from the current sampled set (Fig.~\ref{fig:FPS}), making it unlikely for nearby points to be selected. In large-scale point clouds, many densely clustered points are rarely selected yet still involved in redundant distance computations, causing overhead that scales with input size. To demonstrate this redundancy, we randomly sparsify the input to remove clustered, non-informative points before applying FPS. As shown in Fig.~\ref{fig:pointcloud_distribution_comparison}, even with 50\% sparsification, the spatial distribution after FPS remains well preserved. This reveals the inherent inefficiency of full-cloud computations in FPS and motivates our candidates pruning in \textit{\T}.

\begin{figure}[t]
    \centering
    \begin{subfigure}[t]{0.49\linewidth}
        \centering
        \includegraphics[width=0.65\linewidth]{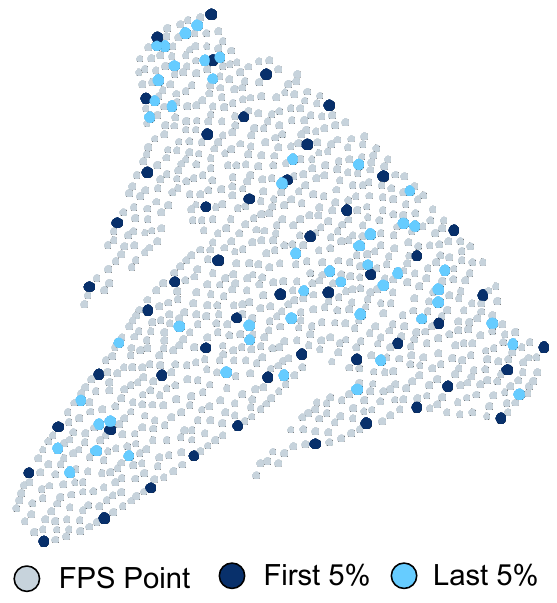}
        \caption{
        Spatial distribution of early (5\%, dark blue) vs. late (5\%, light blue) FPS samples.}
        \label{fig:FPS_point_distribution}
    \end{subfigure}
    \hfill
    \begin{subfigure}[t]{0.49\linewidth}
        \centering
        \includegraphics[width=\linewidth]{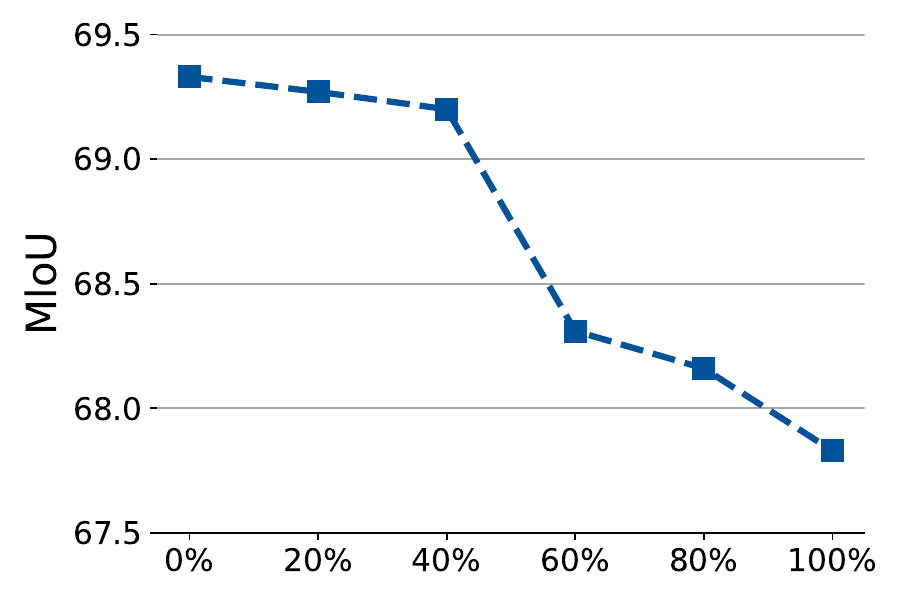}
        \caption{
        PointNeXt-L accuracy after replacing last $p$\% of FPS samples with random points.}
        \label{fig:networkAccuracy_vs_pruneRate}
    \end{subfigure}
    \caption{Analysis of sampling order (a) and iteration redundancy (b) in Farthest Point Sampling (FPS).}
    \label{fig:pointInFPSandAccuracy}
\end{figure}

\noindent\textbf{Redundancy in Iterations.}
FPS prioritizes points farthest from the current sampled set. As shown in Fig.~\ref{fig:FPS_point_distribution}, early-selected points (dark blue) capture the global contour and key structures, while later ones (light blue) mainly refine local details. Since PNNs rely more on early-stage samples for hierarchical feature propagation~\cite{he2025pointrwkv,deng2023pointvector}, later FPS iterations offer diminishing returns. Fig.~\ref{fig:networkAccuracy_vs_pruneRate} confirms this: replacing the final $p$\% of FPS samples with random points has minimal impact on accuracy. These insights motivate our strategy to skip late-stage iterations for faster sampling with negligible accuracy loss.

\begin{figure*}[t]
    \centering
    \includegraphics[width=0.95\linewidth]{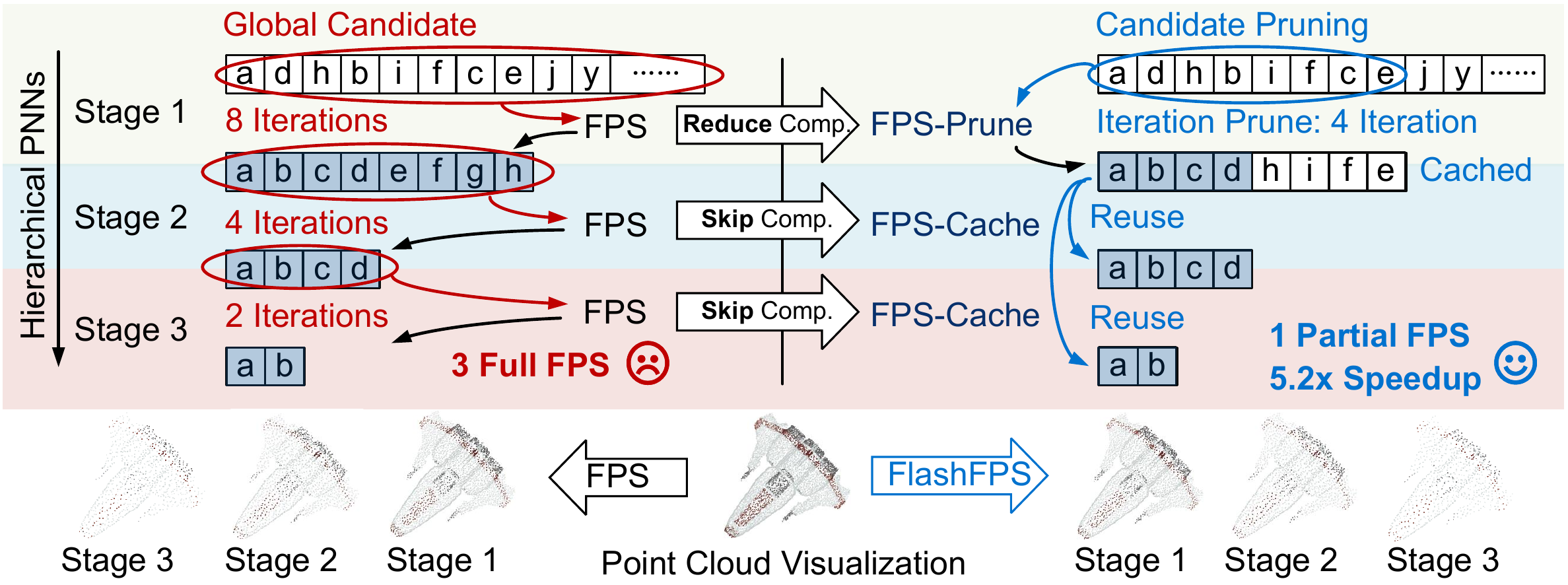}
    \caption{{FlashFPS: a plug-and-play acceleration framework that integrates FPS-Prune (in-layer pruning) and FPS-Cache (cross-layer reuse) to reduce redundancy and enable efficient FPS processing in PNNs.}}
    \label{fig:FlashFPS}
\end{figure*}

Beyond the algorithmic redundancy within FPS itself, additional \textbf{computational redundancy exists at the hierarchical level}. In the original FPS algorithm, the initial seed point is randomly selected. State-of-the-art (SOTA) PNNs typically fix this initial seed point to a default value~\cite{he2025pointrwkv, deng2023pointvector, qian2022pointnext}, making the FPS process fully deterministic for a given input point cloud—the farthest-point order becomes uniquely determined and produces a consistent sequence of sampled points. Under this deterministic setting, we discover that the FPS outputs across layers exhibit a strict \emph{\textbf{prefix property}}: the sampled set of any deeper layer is always a prefix (i.e., subset) of the first-layer output. We formally prove this property and show that reusing the first-layer FPS results incurs \emph{no accuracy loss} (Section~\ref{sec:FPS-Cache}), which directly motivates our \textit{FPS-Cache} design.

To this end, we design \textit{FlashFPS}, a plug-and-play framework tailored to address both algorithmic and hierarchical redundancies in FPS, with three key contributions:
\begin{itemize}[leftmargin=*]
 \item We propose \textbf{FPS-Prune}, a two-level pruning strategy that reduces both the input candidates and the number of iterations to accelerate FPS computation.

\item We propose \textbf{FPS-Cache}, a caching mechanism that eliminates redundant layer-wise recomputation by reusing sampling results across network hierarchies.

\item We provide theoretical insights and extensive empirical results demonstrating that \textbf{the hardware-agnostic \textit{FlashFPS} achieves an average $5.16\times$ speedup on GPUs and $2.69\times$ on PNN accelerators, without any network modification or retraining}.

\end{itemize}

\section{Related Work}

Efforts to accelerate FPS fall into two main categories: GPU-based optimizations and customized hardware designs.

GPU-based FPS works mainly focus on CUDA-level parallelization~\cite{openpoints2023, qi2017pointnet++}. They exploit point-level parallelism within each iteration, processing multiple points simultaneously to update the farthest point. However, they retain full-cloud computation and require complete iterative loops, resulting in high latency for large-scale point clouds despite improved intra-iteration throughput. 
QuickFPS~\cite{han2023quickfps} organizes memory access using a kd-tree and skips partially redundant computations.
However, the additional kd-tree construction introduces extra latency, outweighing the saved FPS time and resulting in an average of 63\% higher latency when the input size is below 30K points.

Customized hardware optimizations employ fine-grained strategies to accelerate FPS~\cite{lin2021pointacc, feng2022crescent, gao2024hgpcn}. Tree-based partitioning methods~\cite{fu2026fractalcloud, zhou202523, zhou2024adjustable} mitigate the full-cloud computation by hierarchically dividing point clouds into blocks and restricting FPS within each block. 
Prediction-based approaches~\cite{yoon2023efficient} estimate sampling distances using one-dimensional features to skip redundant computations. 
However, these methods rely on dedicated hardware support and are poorly suited for GPUs.
For example, block-level partitioning requires distinct and imbalanced data for each FPS instance, preventing data broadcasting and limiting parallel execution, while the irregular and sequential prediction steps in~\cite{yoon2023efficient} further hinder GPU parallelism. 
As a result, these approaches often yield low GPU utilization and may perform worse than CUDA-based FPS.
Beyond limited scalability, these methods substantially modify the standard FPS procedure, necessitating extensive fine-tuning to recover performance and still incurring 1–5\% accuracy degradation in many cases~\cite{zhou2024adjustable, zhou202523, feng2022crescent, kim2021pnnpu}.

These limitations highlight the need for a plug-and-play and hardware-agnostic FPS optimization that generalizes across platforms and PNN architectures.

\section{Proposed Approach}

\subsection{Preliminaries}
\label{sec:prelim}

\paragraph{Farthest Point Sampling (FPS)}

Given an unordered point cloud $\mathcal{P} = \{\mathbf{p}_1, \dots, \mathbf{p}_{N}\}$, where $N$ is the input point size, FPS constructs an ordered subset $\mathcal{P}_1 = [\tilde{\mathbf{p}}_1, \dots, \tilde{\mathbf{p}}_{M_1}]$ via the iterative and greedy rule:
\begin{equation}
\tilde{\mathbf{p}}_{k+1} = \arg\max_{\mathbf{p} \in \mathcal{P} \setminus \mathcal{P}_{1,k}} \min_{\mathbf{q} \in \mathcal{P}_{1,k}} \|\mathbf{p} - \mathbf{q}\|_2,
\end{equation}
where $M_1$ is the sampled point size, $\mathcal{P}_{1,k} = [\tilde{\mathbf{p}}_1, \dots, \tilde{\mathbf{p}}_k]$ is the prefix of the first $k$ sampled points.

\paragraph{Latency Profile of Hierarchical PNNs}
Profiling PointNeXt~\cite{qian2022pointnext} and its successors~\cite{deng2023pointvector} shows that FPS can account for over \textbf{90\%} of total inference time, with the \emph{first} layer alone contributing the lion’s share. The rest of the paper, therefore, focuses on two orthogonal optimizations: \textit{FPS-Prune} — algorithmic pruning inside the first FPS layer, and \textit{FPS-Cache} — architecture-level reuse across layers, with theoretical guarantees ensuring identical sampling correctness.
The overall pipeline of \textit{\T} is illustrated in Fig~\ref{fig:FlashFPS}, and its pseudocode is provided in Algorithm~\ref{alg:flashfps}.

\subsection{FPS-Prune: Intra‑Layer Pruning}

\begin{algorithm}[tb]
\caption{FlashFPS: Accelerated Farthest Point Sampling via Pruning and Cross-Layer Reuse}
\label{alg:flashfps}
\begin{algorithmic}[1]
\footnotesize
\REQUIRE Point cloud $\mathcal{P}$; Point Size $N$; pruning ratio $p \in [0, 1)$
\REQUIRE Sampling sizes $M_1, M_2, \dots, M_L$ \hfill (where $M_L = |\mathcal{P}_L|$, $L$ is the number of layers)
\ENSURE Sampled subsets $\mathcal{P}_1, \mathcal{P}_2, \dots, \mathcal{P}_L$

\STATE \textbf{Layer 1: FPS-Prune with Dual Pruning}
\STATE $\mathcal{P}' \gets$ UniformSample($\mathcal{P}, (1 - p) \cdot N$) \hfill $\triangleright$ Candidate pruning
\STATE $k \gets (1 - p) \cdot M_1$ \hfill $\triangleright$ Iteration pruning
\STATE $\mathcal{P}_{\text{fps}} \gets \textsc{FPS}(\mathcal{P}', k)$
\STATE $\mathcal{P}_{\text{rand}} \gets$ RandomSample $(\mathcal{P}\setminus\mathcal{P}_{\text{fps}}, p \cdot M_1)$
\STATE $\mathcal{P}_1 \gets \mathcal{P}_{\text{fps}} \cup \mathcal{P}_{\text{rand}}$ \hfill $\triangleright$ Sampled points for Layer 1
\STATE $Cache(\mathcal{P}_1)$ \hfill $\triangleright$ Cache sampled points for FPS-Cache

\vspace{8pt}
\STATE \textbf{Layer 2 to $L$: Cross-Layer Reuse via FPS-Cache}
\FOR{$i = 2$ to $L$}
    \STATE $\mathcal{P}_i \gets \mathcal{P}_1[0 : M_i]$ \hfill $\triangleright$ Reuse prefix from cached $\mathcal{P}_1$
\ENDFOR

\vspace{8pt}
\RETURN $\mathcal{P}_1, \mathcal{P}_2, \dots, \mathcal{P}_L$
\end{algorithmic}
\end{algorithm}

\textit{FPS-Prune} is a plug-and-play drop-in that prunes both the input full cloud and iterations while retaining the essential farthest point ordering for key points. Let the pruning ratio be $p \in [0, 1)$. 
$p=0$ is the standard FPS. The algorithm of \textit{FPS-Prune} consists of two steps:

\noindent\textbf{Step 1. Candidate Pruning.}
To mitigate the densely clustered regions, we apply a uniform pre-sampling step that retains only $(1-p)N$ points before running FPS, namely candidate pruning, as described in Line 2 of Algorithm~\ref{alg:flashfps}. 
Because point clouds are inherently unordered and permutation-invariant~\cite{qi2017pointnet++}, uniform pruning produces an unbiased approximation of the original spatial distribution.
Consequently, the reduced input maintains the same overall geometric coverage, and the subsequent distance computations in FPS continue to reflect the statistical representativeness of the retained subset.

\noindent\textbf{Step 2. Iteration Pruning.}
The late-stage iterations of FPS primarily refine local details, which tend to be diminished in deeper PNN layers. Therefore, we apply an iteration pruning, as shown in Lines 3–6 of Algorithm~\ref{alg:flashfps}. FPS now runs for only $(1-p)M_1$ iterations on the pruned set $P'$, generating $\mathcal{P}_{\text{fps}}$ with the first $(1-p)M_1$ critical farthest points that matter most for downstream receptive-field expansion. The remaining $pM_1$ required samples can be filled via random sampling from the input cloud $\mathcal{P}$, excluding the already selected $\mathcal{P}_{\text{fps}}$, which causes negligible accuracy impact ($<0.2\%$) in practice. This process can be efficiently implemented by directly slicing the input, leveraging the unordered nature of point clouds.

This hybrid pruning strategy in \textit{FPS-Prune} mitigates the full-cloud computation in standard FPS, achieving a $\sim1/(1-p)^2$ reduction in pairwise-distance computations. Meanwhile, \textit{FPS-Prune} preserves the crucial “farthest-first” ordering for early-sampled points. As illustrated in Fig.~\ref{fig:pointcloud_distribution_comparison} and Fig.~\ref{fig:pointInFPSandAccuracy}, pruning partial candidates and late iterations maintains the sampling distribution of standard FPS, providing the theoretical intuition for \textit{FPS-Prune}. Empirically, \textit{FPS-Prune} achieves a 3.6× end-to-end speedup on large-scale benchmarks in a training-free manner with negligible overall accuracy loss (less than 0.3\%), making it a practical, plug-and-play alternative for standard FPS.

\subsection{FPS‑Cache: Inter‑Layer Reuse}
\label{sec:FPS-Cache}
\textbf{Motivation: }In hierarchical PNNs, $L$ sampling layers with budgets $M_1 \ge M_2 \ge \dots \ge M_L$ are applied sequentially, where each layer performs FPS on the output of the previous one.
Although every layer launches a fresh FPS pass in the existing implementations~\cite{deng2023pointvector,qian2022pointnext,lin2021pointacc}, the \emph{prefix property} of FPS implies that all FPS operations in deeper layers merely require prefixes of the
first-layer result, which is formally proven later in this section.
Thus, recomputing FPS $L{-}1$ times is redundant. FPS-Cache is proposed to save redundant computation.
\footnote{The idea parallels the \emph{KV-cache} in Transformer inference, where key–value pairs computed for early tokens are reused for later decoding steps.~\cite{kwon2023efficient,zheng2024sglang}}  

FPS-Cache stores the ordered output of the first FPS layer (Line 7 of Algorithm~\ref{alg:flashfps}) and directly reuses its prefixes for subsequent layers according to their required sample sizes (Line 9-11 of Algorithm~\ref{alg:flashfps}), completely eliminating repeated sampling operations.
Since the cached points are exactly those that an independent FPS would have produced, sampling correctness and model accuracy are fully preserved.

As shown in Fig~\ref{fig:FlashFPS}, once the first FPS layer completes, its ordered output is cached and later accessed by deeper layers (e.g., the second and third), fully eliminating redundant FPS computations. This simple yet effective optimization yields a 1.35× end-to-end speedup while incurring negligible memory overhead (3.3 MB for 289K-point inputs with $<0.1\%$ memory overhead), which is negligible compared to the overall model memory footprint.\

\textbf{Theoretical Analysis: }
\label{sec:fpscache_theory}
Let $\mathcal P_1=[\tilde{\mathbf p}_1,\dots,\tilde{\mathbf p}_{M_1}]$ be the first-layer
order obtained by running $M_1$ iterations of FPS on $\mathcal P$ with fixed seed
$\tilde{\mathbf p}_1$. For any $m\le M_1$, write its prefix
$\mathcal P_{1,m}=[\tilde{\mathbf p}_1,\dots,\tilde{\mathbf p}_m]$.
layer $2$ performs an independent FPS run on the candidate set
$\mathcal P_1$, again initialized at $\tilde{\mathbf p}_1$
and using FPS to select $M_2$ points. Similarly,
layer $\ell$ runs on $\mathcal P_{\ell-1}$ to select $M_\ell$ points.

\textbf{Prefix Equivalence: }
We claim that, under the above consistency conditions, running FPS for $m$ iterations
on the restricted candidate set $\mathcal P_1$ returns exactly the prefix $\mathcal P_{1,m}$.
The argument is by induction on $m$. For $m{=}1$, the claim is trivial. Assume the statement holds for $m{=}k$:
the first $k$ selections on $\mathcal P_1$ equal
$\mathcal P_{1,k}=[\tilde{\mathbf p}_1,\dots,\tilde{\mathbf p}_k]$.
At iteration $k{+}1$, standard FPS on the \emph{full} set $\mathcal P$ chooses
\begin{equation}
    \tilde{\mathbf p}_{k+1}
~\in~\arg\max_{\mathbf p\in\mathcal P}\;
\min_{\mathbf q\in\mathcal P_{1,k}}\|\mathbf p-\mathbf q\|_2.
\end{equation}
Because $\mathcal P_1\subseteq \mathcal P$, restricting the maximization to
$\mathcal P_1$ cannot produce a point with \emph{larger} objective value than $\tilde{\mathbf p}_{k+1}$; 
Otherwise, that point would also have been chosen in the full-set run at step $k{+}1$, contradicting the definition of $\tilde{\mathbf p}_{k+1}$. Maximizer within
$\mathcal P_1$ thus equals $\tilde{\mathbf p}_{k+1}$, completing the induction.

\textbf{Hierarchical Reuse: }
Applying the above with $m=M_2$ shows that the layer-2 FPS on $\mathcal P_1$
with seed $\tilde{\mathbf p}_1$ returns $\mathcal P_{1,M_2}$, i.e., the length-$M_2$
prefix of the first-layer order. Repeating this argument on $\mathcal P_2=\mathcal P_{1,M_2}$,
then on $\mathcal P_3=\mathcal P_{1,M_3}$, and so on, yields for every layer $\ell$:
\begin{equation}
\text{Output of layer }\ell \;=\; \mathcal P_{1,M_\ell}.
\end{equation}
Therefore, caching the order $\mathcal P_1$ computed once at layer~1 and handing out
prefixes $\mathcal P_{1,M_\ell}$ to layers $\ell{=}2,\dots,L$ \emph{exactly reproduces}
their independent FPS outputs under the same seed, metric, and tie–breaking rule.
No approximation is introduced, and model accuracy is preserved.

\section{Experiments}
We evaluate \textit{\T} through five main experiments: (1) accuracy and speedup on SOTA PNNs, (2) speedup compared with CUDA-based works across input scales, (3) performance gains on diverse hardware platforms, (4) FPS-Cache memory footprint, and (5) performance in emerging point-cloud LLM scenarios.

\subsection{Experimental Setups} 
We evaluate \textit{\T} on two SOTA hierarchical PNNs, PointNeXt-L~\cite{qian2022pointnext} and PointVector-L~\cite{deng2023pointvector}, using two standard semantic segmentation benchmarks, S3DIS~\cite{armeni20163d} and ScanNet~\cite{dai2017scannet}.
S3DIS comprises six indoor areas (271 rooms, 13 categories) with point counts ranging from 13K–289K, while ScanNet covers 20 indoor categories with 23K–297K points per sample.
We adopt the official open-source configurations and report three standard metrics: overall accuracy \textbf{(OA)}, mean accuracy \textbf{(mAcc)}, and mean Intersection-over-Union \textbf{(mIoU)}, to evaluate the network accuracy.
In addition, \textit{\T} is evaluated on PointLLM~\cite{xu2024pointllm} to assess its applicability in emerging LLM-based point-cloud scenarios.
For GPU speedup, baselines include the standard CUDA-optimized FPS~\cite{openpoints2023} and QuickFPS~\cite{han2023quickfps}.
These experiments are conducted on a workstation equipped with an NVIDIA TITAN RTX GPU (24 GB) and a 48-core Intel Xeon CPU (3.0 GHz).
For cross-platform evaluation, we further test \textit{\T} on two state-of-the-art PNN accelerators, PointAcc~\cite{lin2021pointacc} and Crescent~\cite{feng2022crescent}.
Accelerator results are obtained from a cycle-accurate simulator built on an open-source framework~\cite{guo2023olive} and cross-validated against published performance for fidelity.

\subsection{Evaluation on SOTA PNNs}

Table~\ref{tab:fps_flashfps_comparison} summarizes the evaluation results of \textit{\T} on SOTA PNNs. ``FPS-CUDA'' denotes the baseline using the standard CUDA implementation of farthest point sampling, while “\textit{FlashFPS}-$p$” represents inference with FPS replaced by \textit{\T}, under identical network weights. Here, $p$ indicates the pruning ratio, applied consistently to both candidate pruning and iteration pruning. Speedup is measured by the total inference time across all test samples for each dataset. As expected, a higher pruning ratio yields greater speedup at the cost of slight accuracy degradation. When $p = 25\%$, \T achieves an average $1.62\times$ speedup with less than $0.1\%$ drop across all three metrics (OA, mAcc, and mIoU). Even at an aggressive pruning ratio of $p = 75\%$, \textit{\T} achieves an average $5.16\times$ speedup, while maintaining accuracy losses within $0.3\%$, $0.7\%$, and $0.8\%$ for OA, mAcc, and mIoU, respectively.

These results show that \textit{\T} incurs negligible accuracy loss in SOTA PNNs, well below the commonly accepted $1\%$ threshold~\cite{zhou2024adjustable, zhou2023energy, feng2022crescent, feng2020mesorasi}, while significantly accelerating PNNs inference. This validates \textit{\T} as an effective plug-and-play replacement for FPS, improving efficiency without sacrificing performance.

\begin{table}[tb]
\centering
\small
\caption{{Evaluation of FPS and \textit{\T} for semantic segmentation on S3DIS and ScanNet. \textit{\T-p} integrates FPS-Prune and FPS-Cache, where $p$ denotes the pruning ratio.}}

\scalebox{0.95}{
\renewcommand{\arraystretch}{1}
\begin{tabular}{clcccc}
\toprule[1.5pt]
\makecell{{\textbf{Workload}}} & \textbf{Method} & \makecell{\textbf{OA} \\ (\%)} & \makecell{\textbf{mAcc} \\ (\%)} & \makecell{\textbf{mIoU} \\ (\%)} & \textbf{Speedup} \\
\midrule[1pt]
\multirow{4}{*}{\makecell{\textbf{S3DIS}\\\textbf{PointNeXt-L}}}
        & FPS-CUDA        & 90.09 & 75.76 & 69.33 & 1\\
        & \T-25\%      & 90.05 & 75.7 & 69.21 & \textbf{1.76}\\
        & \T-50\%      & 90.00 & 75.6 & 69.13 & \textbf{2.72}\\
        & \T-75\%      & 89.92 & 75.48 & 69.01 & \textbf{5.28}\\
\cmidrule(l){1-6}
\multirow{4}{*}{\makecell{\textbf{S3DIS}\\\textbf{PointVector-L}}} 
        & FPS-CUDA      & 90.97 & 76.75 & 70.83 & 1\\
        & \T-25\%       & 90.93 & 76.74 & 70.79 & \textbf{1.5}\\
        & \T-50\%       & 90.84 & 76.56 & 70.55 & \textbf{2.76}\\
        & \T-75\%       & 90.7  & 76.32 & 70.27 & \textbf{4.98}\\
\midrule[1pt]
\multirow{4}{*}{\makecell{\textbf{ScanNet}\\\textbf{PointNeXt-L}}} 
        & FPS-CUDA      & 89.07 & 78.66 & 70.54 & 1\\
        & \T-25\%       & 89.07 & 78.62  & 70.45 & \textbf{1.7}\\
        & \T-50\%       & 89.01 & 78.41 & 70.17 & \textbf{2.67}\\
        & \T-75\%       & 88.87 & 78.1 & 69.79 & \textbf{5.27}\\
\cmidrule(l){1-6}
\multirow{4}{*}{\makecell{\textbf{ScanNet}\\\textbf{PointVector-L}}} 
        & FPS-CUDA      & 89.1 & 77.95 & 70.03 & 1\\
        & \T-25\%       & 89.12 & 77.97  & 70.07 & \textbf{1.5}\\
        & \T-50\%       & 89.05 & 77.92 & 69.97 & \textbf{2.8}\\
        & \T-75\%       & 88.78 & 77.24 & 69.27 & \textbf{5.13}\\
\bottomrule[1.5pt]
\end{tabular}
}
\label{tab:fps_flashfps_comparison}
\end{table}


\subsection{Speedup with CUDA-Based works}

\begin{figure}[tb]
    \centering
    \includegraphics[width=\linewidth]{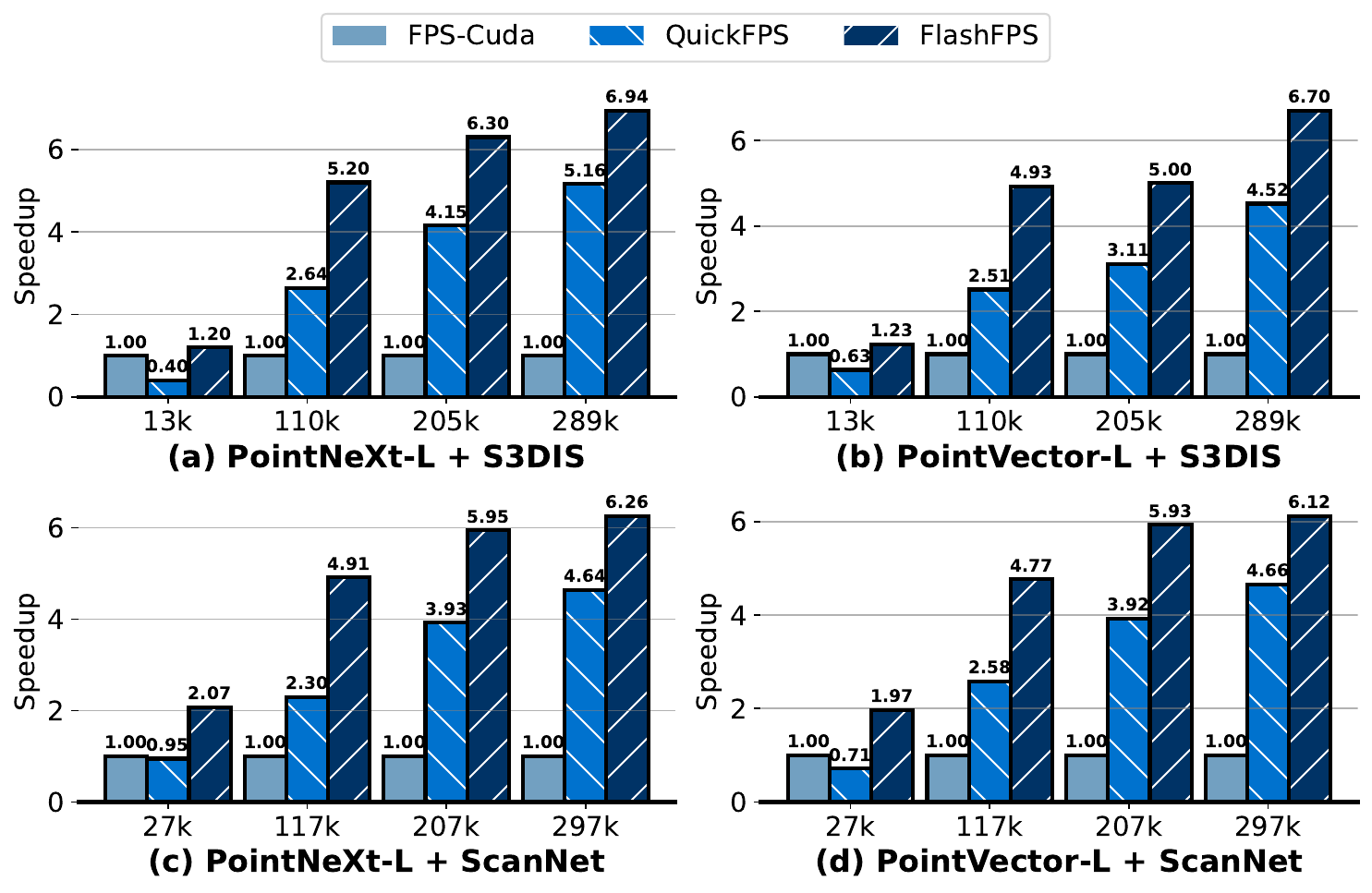}
    \caption{GPU inference speedup of \textit{\T} on PointNeXt-L and PointVector-L for segmenting S3DIS and ScanNet datasets, compared with FPS-CUDA (baseline) and QuickFPS.}
    \label{fig:speedup_breakdown}
\end{figure}

Fig.~\ref{fig:speedup_breakdown} shows the detailed GPU speedup of \textit{FlashFPS} on PointNeXt-L and PointVector-L using the S3DIS and ScanNet datasets under four representative point numbers.
We compare against the CUDA-optimized FPS baseline (FPS-CUDA, from OpenPoints~\cite{openpoints2023}) and QuickFPS~\cite{han2023quickfps}, which is integrated into the same framework for a fair end-to-end comparison.
QuickFPS improves memory access and skips partial redundant computations, with slight accuracy degradation (within 0.5\%). However, its kd-tree construction incurs additional overhead, resulting in an average of 63.4\% longer latency than FPS-CUDA for inputs below 30K points.

In contrast, \textit{\T} exhibits strong scalability, consistently outperforming both FPS-CUDA and QuickFPS across all input scales through its dual-level optimization.
At the algorithmic level, FPS-Prune prunes the input candidates and shortens the iteration loop, yielding average speedups of $3.47\times$ and $3.28\times$ over the baseline for PointNeXt-L and PointVector-L, respectively. At the hierarchical level, FPS-Cache eliminates redundant sampling in deeper network layers, contributing an additional $1.36\times$ and $1.35\times$ speedup for the two models.
Overall, with all workloads involved as in Table~\ref{tab:fps_flashfps_comparison}, \textit{FlashFPS} achieves average $5.16\times$ and $1.96\times$ end-to-end speedup over FPS-CUDA and QuickFPS, respectively, making it particularly effective for scalable 3D perception workloads.

\begin{figure}[tb]
    \centering
    \includegraphics[width=\linewidth]{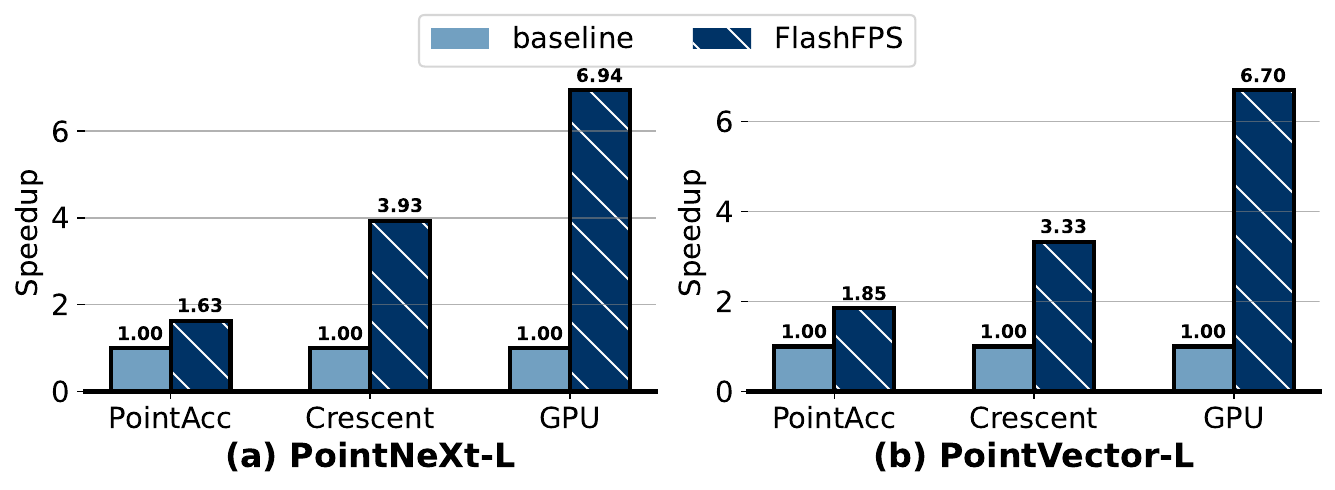}
    \caption{Speedup of FlashFPS on different hardware platforms based on the S3DIS dataset with 289K as input.}
    \label{fig:flashfps_s3dis_backbones}
\end{figure}

\subsection{Speedup on Various Hardware Platforms}

\textit{\T} is designed to be hardware-agnostic and does not rely on any specialized hardware support, making it readily deployable across diverse hardware platforms. We evaluate its effectiveness on a server-class GPU (NVIDIA RTX TITAN) and two SOTA PNN accelerators, PointAcc~\cite{lin2021pointacc} and Crescent~\cite{feng2022crescent}.
Fig.~\ref{fig:flashfps_s3dis_backbones} shows the speedup of \textit{\T} on different hardware platforms based on PointNeXt-L and PointVector-L, with each work’s native optimization as baseline. While existing PNN accelerators focus on parallelism and memory efficiency, they overlook the latency from redundant FPS computations. \textit{\T} addresses this via pruning and caching, reducing both computation and memory access while preserving farthest-selection behavior. Therefore, \textit{\T} achieves additional end-to-end speedups of $1.74\times$ on PointAcc, $3.63\times$ on Crescent, and $6.81\times$ on GPU, demonstrating high efficiency and cross-platform adaptability. On average, \textit{\T} delivers a \textbf{$2.69\times$} speedup on PNN accelerators.

\subsection{Light Memory of FPS-Cache}
\begin{figure}[tb]
    \centering
    \includegraphics[width=\linewidth]{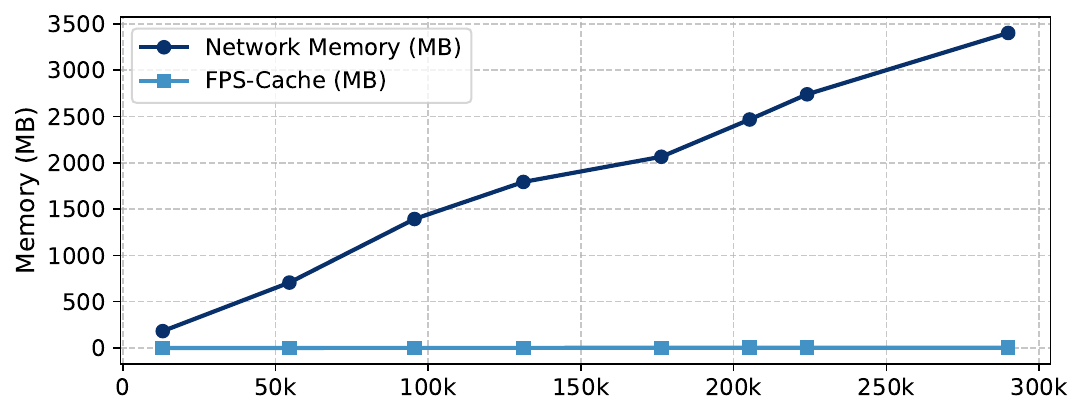}
    \caption{Memory footprint comparison of network inference and FPS-Cache across varying input point scales using PointNeXt-L on the S3DIS dataset.}
    \label{fig:cache_memory}
\end{figure}

FPS-Cache stores the results of the first-layer FPS in PNNs to eliminate redundant computations in subsequent FPS layers, introducing negligible memory overhead. The cached data contains only $N×r$ points, where $N$ is the input size and $r$ is the sampling rate in networks ($r<1$), making the memory requirement grow linearly with input size $N$. 
As shown in Fig.~\ref{fig:cache_memory}, FPS-Cache adds merely {0.15 MB} (0.08\%) memory overhead for 13K points and {3.3 MB} (0.1\%) memory overhead for 289K points, remaining insignificant compared to total network memory. Moreover, the overall network memory grows much faster than the cache overhead, confirming that the additional cost remains negligible even at larger point-cloud scales. Consequently, FPS-Cache achieves an average of 1.35× speedup without accuracy loss, while adding only $\sim$0.1\% extra memory.

\subsection{Speedup for Point-Cloud LLM}


\begin{table}[t]
\centering
\caption{FPS and end-to-end speedups of \textit{FlashFPS} in PointLLM across different input point-cloud sizes.}
\begin{tabular}{c|cc}
\toprule
\textbf{\#Points} & \textbf{FPS Speedup} & \textbf{Overall Speedup} \\
\midrule
16,384  & 15.0$\times$ & 2.10$\times$ \\
32,768  & 15.5$\times$ & 2.90$\times$ \\
65,536  & 15.7$\times$ & 4.56$\times$ \\
\rowcolor{gray!15}
Average  & \textbf{15.4$\times$} & \textbf{3.19$\times$} \\
\bottomrule
\end{tabular}
\label{tab:flashfps_speedup}
\end{table}

\textit{FlashFPS} also accelerates large language models (LLMs) for point cloud understanding.
We integrate it into PointLLM~\cite{xu2024pointllm}, an open-vocabulary 3D reasoning model, where large scenes are first downsampled by FPS into 8K points before being processed by the LLM.
Replacing the standard FPS with \textit{\T} greatly reduces latency.
On the Objaverse~\cite{deitke2023objaverse} dataset using an NVIDIA A100 GPU, \textit{\T} preserves similar geometric structures and identical keyword predictions as standard FPS, while achieving an average $15.4\times$ sampling and $3.19\times$ end-to-end speedup, as listed in Table~\ref{tab:flashfps_speedup}.
These results highlight the efficiency and scalability of \textit{\T} in LLM-based point cloud pipelines.

\subsection{Ablation Study of \textit{FlashFPS} Components} We conduct an ablation study to evaluate the individual and combined effects of the three components in \textit{FlashFPS}: candidate pruning (CP) and iteration pruning (IP) from \textit{FPS-Prune}, and cross-layer reuse via \textit{FPS-Cache} (FC). 
As shown in Table~\ref{tab:ablation} using mIoU as the primary accuracy metric, experiments on PointNeXt-L (S3DIS, 75\% pruning) reveal that all components contribute to speedup, with the full configuration (CP+IP+FC) achieving \textbf{$5.28\times$}.
CP alone yields $2.44\times$ with only 0.32\% accuracy drop, while FC is inherently lossless. 
Notably, network accuracy remains unchanged across CP, CP+IP, and CP+IP+FC, a behavior observed exclusively under the 75\% pruning setting. 
This invariance stems from alignment with the network’s 25\% sampling ratio.
CP prunes 75\% of input points, retaining the 25\% required by the downstream layer, which introduces a slight accuracy loss.
IP then skips 75\% of FPS iterations on this reduced set, leaving the first 25\% of reordered points identical to standard full-loop FPS and therefore not affecting network accuracy.
Finally, FC reuses these sampled points across layers without modification, preserving network accuracy.
These results confirm that all components are effective and that \textit{\T} achieves substantial acceleration without compromising performance.



\begin{table}[tb]
\centering
\renewcommand{\arraystretch}{} 
\caption{Ablation of \textit{\T} components on PointNeXt-L (S3DIS, segmentation task) with 75\% pruning rate. The mIoU is used as the network accuracy metric.}
\begin{tabular}{ccc|cc}
\toprule
w/ CP & w/ IP & w/ FC & mIoU (\%) & Speedup \\
\midrule
\multicolumn{3}{c|}{FPS-CUDA (baseline)} & 69.33 & 1.00 \\
\checkmark &         &         & 69.01 & 2.44 \\
          & \checkmark &       & 69.13 & 2.31 \\
          &         & \checkmark & 69.33 & 1.10 \\
\checkmark & \checkmark &       & 69.01 & 3.63 \\
\checkmark &         & \checkmark & 69.01 & 2.87 \\
          & \checkmark & \checkmark & 69.13 & 2.77 \\
\rowcolor{gray!15}
\checkmark & \checkmark & \checkmark & \textbf{69.01} & \textbf{5.28} \\
\bottomrule
\end{tabular}
\label{tab:ablation}
\vspace{-4pt} 
\end{table}

\section{Conclusion} 
We propose \textit{\T}, a plug-and-play acceleration framework for farthest point sampling in point-based neural networks. \textit{FlashFPS} incorporates two complementary optimizations: \textit{FPS-Prune}, which accelerates FPS through pruning, and \textit{FPS-Cache}, which removes redundant operations across network layers. Experiments demonstrate that \textit{\T} achieves average end-to-end speedups of {5.16$\times$} on GPUs and {2.69$\times$} on PNN accelerators, while maintaining network accuracy, enabling efficient and scalable PNN inference.


\begin{acks}
This work was supported in part by NSF grants 2112562 and by ARO W911NF-23-2-0224. The authors would like to thank the anonymous reviewers.
\end{acks}

\newpage

\bibliographystyle{ACM-Reference-Format}
\bibliography{ref}

\end{document}